\documentclass[mlmain]{jmlr}
\jmlrpmlr
\usepackage[T1]{fontenc}
\usepackage{lmodern}
\usepackage{xurl}
\usepackage{booktabs}
\newcommand{\doi}[1]{doi: \begingroup\urlstyle{rm}\def\UrlBreaks{\do\/\do\-\do\.}\href{https://doi.org/#1}{\color{black}\nolinkurl{#1}}\endgroup}

\jmlrvolume{}
\firstpageno{1}
\editors{}
\jmlryear{2026}
\jmlrworkshop{Symmetry and Geometry in Neural Representations}

\author{\Name{I\~nigo Parra} \Email{iparra@berkeley.edu} \\
       \addr Department of Linguistics \\ University of California, Berkeley, USA}

\title[Natural-language Training Shapes Brain-aligned Geometry]{Language, Not Code Abstraction: Natural-language Training Shapes Brain-aligned Geometry}

\hypersetup{pdftitle={Language, Not Code Abstraction: Natural-language Training Shapes Brain-aligned Geometry}, pdfauthor={I\~nigo Parra}, pdfkeywords={representational geometry; language-brain alignment; CKA; RSA}}

\begin{document}
\maketitle

\begin{abstract}
Code-trained language models produce positive brain scores, which raises the question of whether abstract structure learned from programming languages can substitute for natural-language training. We trained 25 GPT-2-small models on TinyStories, Python, C functions, x86-64 assembly, and the corresponding machine code, and scored them using a shared tokenizer and encoding procedure against matched random-weight models. On Pereira-243 and Tuckute2024, TinyStories exceeds the mean of the four code models by $\Delta r = 0.053$ and $0.059$ (Holm-corrected $p<0.001$), while code training leaves alignment near the random-weight level. The language advantage is benchmark-specific: Blank2014 shows no corrected language-code difference, and Fedorenko2016 shows no language benefit. Geometry tracks the encoding result on Pereira-243: TinyStories seeds converge on the most similar representations, and its fit-free correspondence with neural dissimilarity peaks at $\rho = 0.14$, twice the code and random-weight levels. Binary-trained models have about three effective dimensions, so seed agreement alone does not imply neural correspondence. Updating only the embedding matrix on English improves mean-over-depth alignment for every code model more than an equal update on its own corpus. Within this architecture and training setup, natural-language training produces a distinct representational geometry with closer neural correspondence on Pereira-243.\footnote{Code and results: \url{https://github.com/IParraMartin/language-training-brain-geometry}. \\ Models: \url{https://huggingface.co/collections/IParraMartin/brain-scores}.}
\end{abstract}

\begin{keywords}
representational geometry; language-brain alignment; centered kernel alignment; representational similarity analysis; code language models.\\
\noindent \textbf{Project page}: \href{https://iparramartin.github.io/language-training-brain-geometry/}{\texttt{language-training-brain-geometry}}
\end{keywords}

\section{Introduction}
\label{sec:intro}

Transformer language models predict activity in the human language network \citep{schrimpf2021neural,caucheteux2022brains,goldstein2022shared}. Models trained on computer code also obtain positive scores, suggesting alignment might arise from abstract sequential and compositional structure rather than from learning natural language. An absolute score cannot settle this. A randomly initialized transformer already supplies a learned tokenizer, positional structure, causal attention, nonlinear random features, and, under encoding analyses, a fitted readout \citep{pasquiou2022notborn,alkhamissi2024shallow}. Surface properties of the stimuli add predictive power \citep{feghhi2024brainscores}. A corpus contributes what it produces relative to an identical model that was never trained.

We contrast two predictions. If abstraction learned from code drives alignment, models trained on high-level source code should align better than those trained on lower-level representations, giving a Python $>$ C $>$ assembly $>$ binary ordering. If natural-language learning drives it, a model trained on English should separate from all code models, with no particular ordering among them. Our five corpora share one tokenizer, architecture, and training recipe. Three of them keep the same programs fixed: every C function in the corpus also appears as x86-64 assembly and as hexadecimal machine code. Python is an unrelated corpus, and TinyStories supplies narrative English \citep{eldan2023tinystories}.

We score every block of every model on Pereira2018 \citep{pereira2018universal}, Tuckute2024 \citep{tuckute2024driving}, Blank2014 \citep{blank2014dissociation}, and Fedorenko2016 \citep{fedorenko2016ecog}, treating independently trained seeds as the unit of inference. Because fitted encoders can exploit stimulus surface features, we add analyses that fit nothing: CKA between models, covariance spectra, and RSA between model and brain. Two interventions—block transplants and embedding-only retraining—ask which learned components carry the effect. The results support a narrower claim than the one we began with: English training produces a depth-wide encoding advantage over code on Pereira-243 and Tuckute2024, but not on Blank2014 or Fedorenko2016. On Pereira-243, it also produces a more reproducible geometry with closer neural correspondence. No abstraction ordering appears among the code models. Because the models are compared with human responses to language, these experiments constrain what code training contributes to language-network alignment, not how people process code, which recruits mainly domain-general cortex \citep{ivanova2020codecomprehension,liu2020codelogic}.

\section{Methods}
\label{sec:methods}

\subsection{Training corpora}
\label{sec:corpora}

TinyStories is the original training split \citep{eldan2023tinystories}. The C corpus contains 2.8M functions, and the assembly and binary corpora contain the same functions row by row: C function names reappear in the aligned assembly in 1,992 of 1,993 sampled rows, and the hexadecimal bytes decode to the same instructions. Binary is therefore text, not a raw byte stream: two-digit hexadecimal tokens for each function's unlinked x86-64 machine code, with a few undocumented annotation tokens. Python comprises 165k files from an unidentified source, 64\% of which are exact duplicates; parsing and regenerating each file removed comments, but identifiers and string literals (23\% of characters) remain, and about 6\% of C functions retain English comments. We do not record the compiler or its flags, and we do not record the upstream code collections.

\subsection{Models and training}
\label{sec:training}

A 50k-token byte-level BPE tokenizer, trained once on a shuffled 50MB sample of each corpus, segments all corpora and all neural stimuli. Every model is GPT-2 small \citep{radford2019gpt2} (12 blocks, width 768, context 1,024), trained with AdamW \citep{loshchilov2019adamw} (learning rate $10^{-4}$, weight decay 0.01, 1,000 warmup updates, cosine decay, bf16) on batches of $16\times1{,}024$ tokens for at most 30,517 updates. Documents are truncated to 1,024 tokens and packed; 10\% is held out to select the best checkpoint (every 1,000 updates, patience 5). Seeds 42-46 produce five models per corpus; seed 42 splits packed sequences and inserts no document separators, whereas seeds 43-46 split documents before packing with an end-of-document token.

The recipe is shared, but realized exposure is not. Every logged run (seeds 43-46) reached the update cap and was evaluated at update 30,000; seed 42's logs were not retained. After truncation, that budget is 6.8 passes over the Python training split, 1.2 over TinyStories, 1.0 over C, 0.61 over binary, and 0.35 over assembly (Table~\ref{tab:exposure}). The design therefore fixes architecture, tokenizer, recipe, and maximum budget, not repetition, unique content, or optimization difficulty. Three random-weight models with the same tokenizer, never updated, provide a matched reference for what training adds; they are not a chance level.

\subsection{Neural benchmarks and encoding}
\label{sec:benchmarks}

We use the Brain-Score language implementation \citep{schrimpf2020integrative} at commit \texttt{6650678} (Appendix~\ref{app:benchmarks}). Pereira-243 and Pereira-384 are sentence sets from two experiments of one fMRI reading study; Tuckute2024 is one language-network response averaged over the five participants who read 1,000 sentences; Blank2014 has fMRI time points from five listeners to eight stories; Fedorenko2016 has word-by-word ECoG from five patients. For each stimulus, we extract the final token's activation in each block after the model reads the cumulative passage, story, or sentence context. We fit ridge regressions to predict neural responses, selecting the penalty within each training fold. Folds are grouped by context except for random sentence splits on Tuckute2024. We report raw Pearson $r$ on held-out data, averaged over neuroids within folds and then over folds, and compare conditions only within a benchmark.

\subsection{Statistical inference}
\label{sec:stats}

We use a model's prespecified summary, its mean $r$ over all twelve blocks, so we don't select a favorable layer after seeing scores. On each of the four benchmark families, we test TinyStories minus random weights, the four-code mean minus random weights, and TinyStories minus the code mean, as weighted differences of independent group means with Welch-Satterthwaite degrees of freedom; the three random initializations enter as a sampled group, and Holm correction covers the twelve tests \citep{holm1979simple}. Block-wise tests compare five TinyStories with 20 code models (Welch) and test heterogeneity among code corpora (ANOVA), each corrected over blocks. These tests use seeds as replicates \citep{nadeau2003inference,bengio2004nounbiased,varoquaux2017decoders}.

\subsection{Representational geometry}
\label{sec:methods-geometry}

Geometry uses the same Pereira-243 stimuli and cumulative contexts as encoding. Hidden states are taken after each block, including final layer normalization at block 11. Linear CKA \citep{kornblith2019similarity} measures how similarly two models arrange the stimuli up to rotation and scale: between seeds, it measures reproducibility; between corpora, distinctness. The participation ratio $(\sum_i\lambda_i)^2/\sum_i\lambda_i^2$, where $\lambda_i$ are covariance eigenvalues, estimates how many directions carry substantial variance. RSA \citep{kriegeskorte2008rsa} correlates (Spearman $\rho$) the upper triangles of model and brain dissimilarity matrices (one minus Pearson correlation across features); it does not fit a model-to-brain map, although it remains sensitive to whatever structure dominates both matrices. We remove leading principal components before RSA to test whether a few dominant axes carry it.

\subsection{Controls and interventions}
\label{sec:methods-interventions}

We measure how many stimulus words each corpus contains, how often the stimulus tokens occur, and seed-42 BLiMP accuracy \citep{warstadt2020blimp}. A form control retrains on TinyStories after a one-to-one pseudoword relabeling of content words, and a $2\times2$ design crosses word-order shuffling with a permutation of words within part-of-speech classes (three seeds per cell). Progressive transplants copy the first $k$ TinyStories blocks into a code receiver, with or without the TinyStories token and position embeddings, reading at block 9. Embedding-only adaptation updates just the token embedding matrix (tied to the output layer) on TinyStories or on the receiver's own corpus with equal budgets. Without a learned stitch \citep{bansal2021stitching}, a failed transplant cannot distinguish representational incompatibility from missing transferable information.

\begin{figure}[t]
  \centering
  \includegraphics[width=\textwidth]{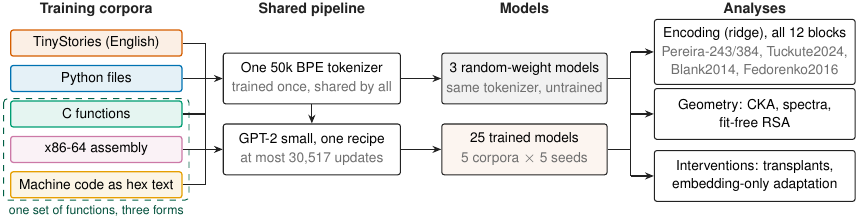}
  \caption{Design. Five corpora share one tokenizer and one GPT-2-small recipe, yielding 25 trained models (five seeds per corpus); C, assembly, and binary share the same functions across three representations. The matched reference comprises three random-weight models for encoding and two for geometry. Every model is read at all twelve blocks on four benchmark families, with Pereira contributing two stimulus sets.}
  \label{fig:design}
\end{figure}

\section{Results}
\label{sec:results}

\subsection{English training adds alignment on two of four benchmarks}
\label{sec:results-gain}

Random weights already score above zero, so a positive code score does not show that code training helped. Table~\ref{tab:benchmarks} gives mean-over-depth scores and the prespecified contrasts. On Pereira-243 and Tuckute2024, TinyStories exceeds the code mean by $\Delta r = 0.053$ (95\% CI $[0.046, 0.060]$) and $0.059$ ($[0.043, 0.074]$), both surviving correction, while the code mean stays near random weights ($+0.004$ and $-0.007$). TinyStories exceeds random weights by $0.057$ and $0.051$, and every TinyStories seed exceeds every random initialization on both benchmarks (exact rank test, $p = 0.036$), but with three initializations these contrasts have only 2.3 and 3.4 degrees of freedom and do not survive correction (Holm $p = 0.145$).

Blank2014's TinyStories and code gains over random weights ($+0.014$, $+0.009$) yield no corrected language-code difference ($+0.005$). On Fedorenko2016, both means round to $r = 0.170$, below random weights.

\begin{table}[t]
  \centering
  \caption{Mean-over-depth encoding scores (raw Pearson $r$ averaged over the twelve blocks, five seeds per trained condition, three random initializations) and the three prespecified contrasts. ``Code'' is the equally weighted mean of Python, C, assembly, and binary. Bold contrasts survive Holm correction over all twelve tests, with random-initialization variance included. Intervals and $p$-values are in Table~\ref{tab:primary-tests}.}
  \label{tab:benchmarks}
  \small
  \setlength{\tabcolsep}{5pt}
  \begin{tabular}{lrrrrrr}
\toprule
Benchmark & Random & Code & TinyStories & TS$-$rand. & Code$-$rand. & TS$-$code \\
\midrule
Pereira-243 & 0.064 & 0.069 & 0.122 & +0.057 & +0.004 & \textbf{+0.053} \\
Tuckute2024 & 0.274 & 0.266 & 0.325 & +0.051 & $-$0.007 & \textbf{+0.059} \\
Blank2014 & 0.040 & 0.049 & 0.054 & +0.014 & +0.009 & +0.005 \\
Fedorenko2016 & 0.175 & 0.170 & 0.170 & $-$0.004 & $-$0.005 & +0.000 \\
\bottomrule
\end{tabular}
\end{table}

\subsection{The language-code gap spans depth}
\label{sec:results-corpus}

\begin{figure}[t]
  \centering
  \includegraphics[width=\textwidth]{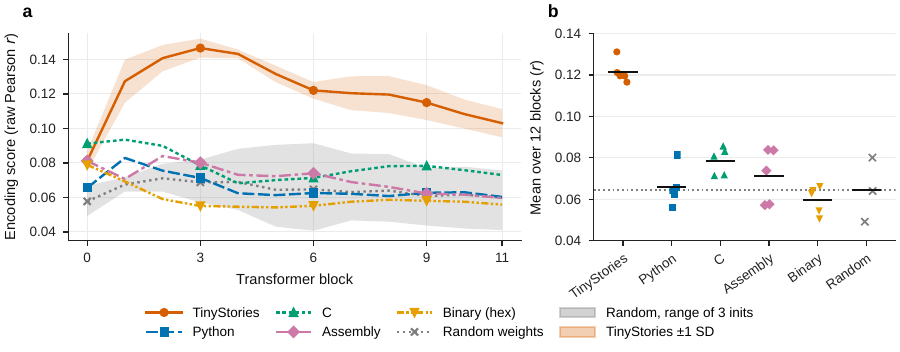}
  \caption{Pereira-243. \textbf{(a)} Encoding score at each block: mean of five seeds per corpus, with $\pm1$ SD across seeds for TinyStories; the gray band spans the three random initializations and the dotted line is their mean. \textbf{(b)} Mean over twelve blocks for each model (points; bars are condition means). The dotted line is the mean. Seeds are the replicates; folds do not enter the uncertainty.}
  \label{fig:pereira}
\end{figure}

A depth mean could mask a gap confined to a single favorable layer. On Pereira-243 (Figure~\ref{fig:pereira}), TinyStories and code models do not differ significantly at block 0 ($p_{\text{Holm}} = 0.61$), and TinyStories is significantly higher at every later block, with a selected maximum gap of $\Delta r = 0.079$ at block 4. Pereira-384 and Tuckute2024 show the same pattern at 11 of 12 blocks, again sparing block 0 (Appendix~\ref{app:primary}). The gap is a property of most of the stack rather than of one readout; the profiles show where it is expressed, not which computation produces it.

\subsection{English training produces a reproducible, brain-like geometry}
\label{sec:results-geometry}

\begin{figure}[t]
  \centering
  \includegraphics[width=\textwidth]{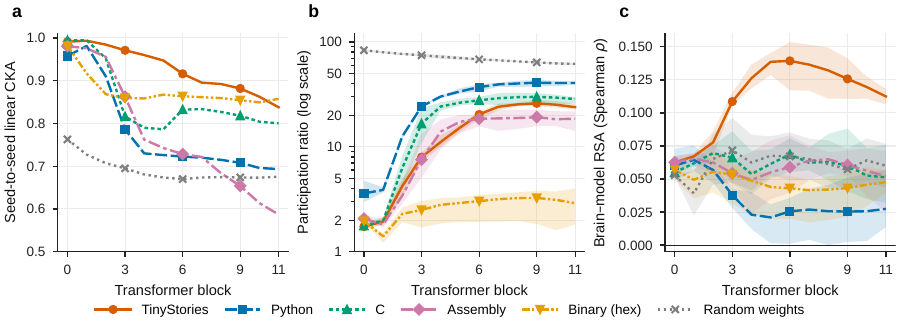}
  \caption{Representational geometry on Pereira-243 stimuli (cumulative topic context, final token). \textbf{(a)} Mean linear CKA between seeds of the same corpus. \textbf{(b)} Participation ratio, log scale; bands span the seeds. \textbf{(c)} Brain-model RSA (Spearman $\rho$ between dissimilarity matrices); lines are seed means and bands span the seeds. Five seeds per corpus and two random initializations; all panels are descriptive.}
  \label{fig:geometry}
\end{figure}

Encoding scores depend on a fitted readout, so we asked whether the representations themselves differ (Figure~\ref{fig:geometry}). Across blocks 4-11, TinyStories seeds have the highest mutual CKA (0.90), then binary (0.86), C (0.81), Python (0.71), assembly (0.69), and random weights (0.67). CKA among TinyStories seeds exceeds that between TinyStories and code models by 0.06 at block 0 and 0.39 from block 4 on (Table~\ref{tab:cka-depth}). Training compresses every corpus relative to random weights (participation ratio 83 at block 0 and 62 at block 11): binary occupies about two to three effective dimensions across blocks, whereas TinyStories expands to about 26.

These properties come apart. Binary is nearly as reproducible as TinyStories while occupying only about three effective dimensions, so seed agreement alone does not identify a brain-like representation. Fit-free RSA separates the corpora: TinyStories rises from $\rho = 0.06$ at block 0 to $0.14$ at block 6, while every code model peaks at or below $0.07$, the random-weight range, and Python falls below random weights in later blocks. Removing leading principal components preserves the TinyStories lead (Table~\ref{tab:pc-ablation}). The RSA advantage is specific to Pereira-243: on Pereira-384 and Fedorenko2016, binary reaches the highest RSA, and we do not interpret RSA for Tuckute2024's single averaged response or Blank2014's truncated contexts (Appendix~\ref{app:geometry}).

\subsection{Code representations differ, but not along an abstraction ordering}
\label{sec:results-hetero}

The code models are not interchangeable. Their encoding scores differ significantly at 2 of 12 Pereira-243 blocks and 8 of 12 Pereira-384 blocks, and their geometries differ more: at blocks 4-11, Python-C CKA is 0.50, below the within-corpus agreement of either (0.71, 0.81); Python is the most distinct code representation (0.23 with binary), and C and assembly, which hold the same functions, are the most similar pair (0.62).

These differences do not follow the predicted ordering. C leads the code models at 8 of 12 Pereira-243 and all 12 Pereira-384 blocks, assembly at the other four Pereira-243 blocks, and Python never leads; the full Python $>$ C $>$ assembly $>$ binary order occurs in none of the 60 block profiles across the five benchmark sets. Within the matched triplet, the Pereira depth summaries do decline from C to assembly to binary (0.079, 0.071, 0.060 on Pereira-243), so on Pereira the ordering fails because Python, the highest-level language, scores below C; on the other three benchmarks binary leads the code models. Two cautions apply: the triplet holds programs fixed, but the models saw 1.0, 0.35, and 0.61 passes of them, and Python differs from C in content, selection, and level. The result argues against this abstraction hierarchy under these conditions, not against any contribution of abstraction or code learning to alignment.

\subsection{The advantage persists after content-word relabeling}
\label{sec:results-confounds}

English training also means seeing the stimulus words, but observational diagnostics suggest that lexical exposure alone is insufficient (Appendix~\ref{app:confounds}). Binary contains 1\% of the Pereira word types and Python 98\%, yet both stay near the random-weight level; the average stimulus token is more frequent in assembly (6.3M occurrences) than in TinyStories (4.1M). A five-corpus correlation ($r = 0.38$, $p = 0.53$) does not rule out a relationship, so these counterexamples support the argument. BLiMP separates TinyStories (0.64) from the code models (0.53-0.55), but orders the code models differently from the neural scores.

The form control is stronger. Relabeling every TinyStories content word with a pseudoword leaves alignment comparable at 500M tokens (mean of blocks 0, 6 and 11: 0.107 intact, 0.115 relabeled; three seeds, Welch $p = 0.12$). The one-to-one map preserves word types, frequencies, co-occurrences, and syntactic frames and changes only the forms; a comparable endpoint is not an equivalence test. In the $2\times2$ design, neither manipulation produces a statistically detectable decrease at 500M tokens (intact minus shuffled: $+0.004$, $p = 0.38$; intact minus permuted: $+0.007$, $p = 0.14$), although combining both gives the lowest cell (0.096). The pseudoword model therefore retains substantial alignment without familiar content-word forms; whether preserved regularities suffice remains open.

\subsection{The embedding-block interface constrains transfer}
\label{sec:results-locus}

\begin{figure}[t]
  \centering
  \includegraphics[width=\textwidth]{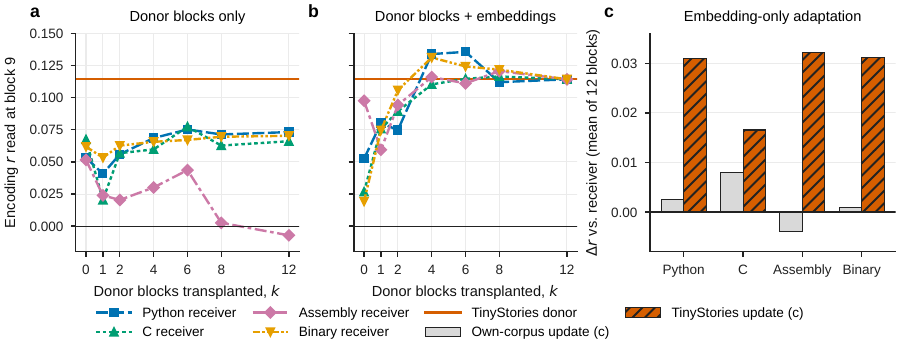}
  \caption{Interventions on Pereira-243, one donor (TinyStories, seed 42) and one receiver checkpoint per code corpus. \textbf{(a)} Encoding $r$ at block 9 after replacing the receiver's first $k$ blocks with the donor's; $k = 0$ is the untouched receiver and $k = 12$ also copies the final normalization. \textbf{(b)} The same block transplants, but also copies the donor's token and position embeddings; $k = 12$ is the donor. \textbf{(c)} Change in mean-over-depth $r$ after updating only the tied embedding matrix on the receiver's own corpus or on TinyStories (equal budgets).}
  \label{fig:locus}
\end{figure}

Transplanting TinyStories blocks alone does not reproduce the donor's alignment (Figure~\ref{fig:locus}a). With all twelve blocks over code embeddings, the Python, C, and binary receivers stay near $r = 0.07$ at block 9, and assembly falls to $-0.01$. Adding the TinyStories token and position embeddings (Figure~\ref{fig:locus}b), every receiver reaches the donor's level ($r = 0.114$) by $k = 4$ and equals it at $k = 12$, where the model is the donor. Retraining only the tied embedding matrix on TinyStories raises the mean-over-depth score of every code model ($+0.028$ on average), compared with a smaller mean change under the same budget on its own corpus ($+0.002$; paired $t(3) = 4.31$, $p = 0.023$; Table~\ref{tab:embed-adapt}). This paired comparison describes four fixed checkpoints, rather than generalizing over code languages. English embeddings alone do not reproduce the donor. Compatibility with downstream blocks therefore matters, and a learned stitch might enable block-only transfer; the effect is not localized to embeddings.

\subsection{Benchmark-specific outcomes constrain interpretation}
\label{sec:results-benchmarks}

Blank2014's story contexts exceed the 1,024-token window for 417 of 1,317 time points. Encoding keeps the most recent tokens, but our RSA extraction keeps the first 1,024, so 411 time points have an input identical to an earlier one; its near-zero RSA says nothing about the neural data. On Fedorenko2016, neither TinyStories nor the pooled code mean significantly exceeds random weights on word-by-word ECoG responses, and binary has the highest descriptive mean. A model is not simply more brain-like: the dataset, readout, and reference determine which differences an analysis can detect.

\section{Discussion}
\label{sec:discussion}

On Pereira-243 and Tuckute2024, English-trained models exceed code models in encoding at almost every block. On Pereira-243, they also show a more reproducible geometry with closer correspondence to neural dissimilarity. Code training keeps scores near the random-weight reference, so code models' positive scores largely reflect what an untrained network of the same architecture achieves through its tokenizer, random features, and fitted readout. This motivates interpreting scores as increments over a matched reference.

The geometry separates three easily conflated properties: reproducibility across seeds, dimensionality, and neural correspondence. Binary models agree almost as well as English models with about three effective dimensions and no neural advantage; Python models spread variance widely, agree less consistently, and drift below the random-weight RSA. On Pereira-243, only English training combines high reproducibility, intermediate dimensionality, and the highest fit-free correspondence. We read this as evidence that the training distribution organizes representational geometry in a way the brain comparison detects, not as a mechanism shared with people. The interventions suggest where that organization becomes usable: English blocks need compatible embeddings, and English statistics written into a code model's embeddings help more than further code training. One hypothesis, consistent with the pseudoword control, is that English training shapes the input space so downstream blocks can compose distributional regularities that also structure the brain's responses to sentences; testing it requires stitching and manipulations of distributional structure rather than word forms.

Code models do learn different geometries: Python and C occupy distinct spaces on the same stimuli, and the matched C and assembly models are closest. Brain alignment does not follow the abstraction level that separates them. Because the activations are elicited by English sentences rather than native code, and Python differs from C in content as well as level, we do not attribute these geometric differences to properties of particular programming languages.

\section{Conclusion}
\label{sec:conclusion}

With the architecture, tokenizer, recipe, and encoding procedure shared, training on narrative English produces a depth-wide alignment advantage over code on Pereira-243 and Tuckute2024, along with a reproducible geometry that more closely matches neural dissimilarity on Pereira-243. Code training stays near a matched random-weight reference, and the proposed Python $>$ C $>$ assembly $>$ binary ranking does not appear. The advantage is absent on Blank2014 and Fedorenko2016, depends on compatibility between embeddings and blocks, and persists after content-word relabeling. Brain scores are most informative as increments over matched references, reported with fit-free geometry and benchmark-specific outcomes.

\paragraph{Limitations.}
We study one GPT-2-small architecture, one English distribution, and one tokenizer trained on all five corpora. Exposure ranged from 0.35 passes over assembly to 6.8 over a heavily duplicated Python corpus, whose validation documents largely recur in training. Inference generalizes over optimization runs of this design, not over participants, stimuli, architectures, or training distributions; the random reference has three initializations, and transplants use one checkpoint pair per code corpus without a learned stitch.

\clearpage
\bibliography{references}

\clearpage
\appendix

\section{Training corpora and realized exposure}
\label{app:corpora}

\paragraph{Sources and representation.} The five corpora are configurations of one public dataset \citep{parra2026codeenglish} (revision \texttt{cc83566}, last modified 16 May 2026, before the first training run). Table~\ref{tab:provenance} reports what we measured on the stored files with \path{scripts/reproduction/audit_corpora.py}. TinyStories is the original training split \citep{eldan2023tinystories}: its 2,119,719 rows match the source release, and the first 2,000 rows are identical. The C, assembly, and binary configurations share the same 2.8M row identifiers and hold one function per row in three forms. The assembly is x86-64 AT\&T syntax with \texttt{.cfi} directives, \texttt{endbr64} instructions, and frame-pointer prologues with arguments spilled to the stack, consistent with unoptimized compilation; the compiler, its version, and its flags are not recorded. The binary form writes each function's machine code as space-separated two-digit hexadecimal tokens: call and RIP-relative displacements are zero, as in unlinked object code, and a leading token and single tokens after branch and relocation fields are not part of the instruction stream; their encoding is undocumented. For example, the first row's C function \texttt{timespec\_get} appears as \texttt{endbr64; pushq \%rbp; movq \%rsp,\%rbp; \ldots} in the assembly configuration and as \texttt{64 f3 0f 1e fa 55 48 89 e5 \ldots} in the binary configuration. Python files are normalized by parsing and regenerating them (99.98\% of sampled files are fixed points of this operation), which removes comments; string literals, identifiers, and an author line such as \texttt{\_\_author\_\_} remain. C functions keep their comments in about 6\% of cases. Assembly and binary contain no source comments because they are compiler output, although symbol names derived from the C identifiers remain in the assembly. The dataset and training code do not reveal the upstream collections, selection criteria, or deduplication of the C and Python data.

\paragraph{Duplication and validation overlap.} Python has 59,456 distinct files among 165,440 rows, and 72.8\% of the validation documents of a seed-43 split also occur in its training split, so Python's validation loss overstates generalization. TinyStories has 15\% duplicate rows, binary 27\% (many distinct C functions compile to identical machine code), assembly 4\% and C none. Splits are random over rows, with no grouping by repository, program, or compilation; the matched C-assembly-binary rows are split independently in each corpus.

\paragraph{Tokenizer.} The shared tokenizer was trained once on a shuffled 50MB sample from each of the five corpora (the training script's byte budget), with a 50,000-entry vocabulary and one special token; all checkpoints carry the identical vocabulary and merges.

\begin{table}[htbp]
  \centering
  \caption{Corpus provenance measured on dataset revision \texttt{cc83566}. ``\% val.\ in train'': share of validation documents (seed-43 split) whose exact text also appears in the training documents. ``\% truncated'': documents longer than the 1,024-token limit, which training cuts to their first 1,024 tokens.}
  \label{tab:provenance}
  \scriptsize
  \setlength{\tabcolsep}{3pt}
  \begin{tabular}{lllrrrr}
\toprule
Corpus & Unit & Stored representation & Rows & \% unique & \% val.\ in train & \% truncated \\
\midrule
TinyStories & story & English prose & 2.1M & 85 & 27.0 & 0.0 \\
Python & file & Python, AST-normalized & 165k & 36 & 72.8 & 24.1 \\
C & C function & C source & 2.8M & 100 & 0.0 & 2.4 \\
Assembly & function & x86-64 AT\&T assembly & 2.8M & 96 & 6.7 & 20.7 \\
Binary (hex) & function & hex text of machine code & 2.8M & 73 & 32.1 & 8.1 \\
\bottomrule
\end{tabular}
\end{table}

\paragraph{Packing and exposure.} Each document is tokenized and truncated to 1,024 tokens, an end-of-document token is appended (seeds 43-46 only), and consecutive batches of 1,000 documents are concatenated and cut into 1,024-token sequences, dropping each batch's remainder; sequences contain no padding. Seed 42 packed the whole corpus and then held out 10\% of the packed sequences; seeds 43-46 held out 10\% of documents and packed each split separately. The end-of-document token is also the padding token, so the language-modeling loss ignores it. We recomputed the number of training sequences for every run by replaying these steps on the stored corpus (\path{scripts/reproduction/recover_packed_splits.py}). Each update processes $16\times1{,}024 = 16{,}384$ input positions. For the logged runs (seeds 43-46), every run stopped at the 30,517-update cap without triggering early stopping; the selected checkpoint is update 30,000 (491.5M positions), and best-validation weights were restored at the end. Pass equivalents divide processed positions by the training-split size; they are not counts of distinct documents because the sampler reshuffles after each pass and counts duplicates separately. Seed 42's step count and selected checkpoint are not recorded; its pass equivalents assume the same checkpoint. The complete per-run report, including fingerprints, hyperparameters, and missing values, is \texttt{training\_exposure.csv}, written by \texttt{make results} in the code repository.

\begin{table}[htbp]
  \centering
  \caption{Realized training exposure. Tokens kept: corpus size after per-document truncation. Training sequences: range over the five seeds. Passes: processed tokens at the evaluated checkpoint divided by the training-split size, for the four logged runs. The full-split token counts in Table~\ref{tab:corpora} overstate the available data for every corpus except TinyStories.}
  \label{tab:exposure}
  \scriptsize
  \setlength{\tabcolsep}{3pt}
  \begin{tabular}{lrrrrrr}
\toprule
Corpus & Documents & \% truncated & Tokens kept & Train sequences & Logged runs & Passes (logged) \\
\midrule
TinyStories & 2.1M & 0.0 & 463M & 406.1-408.0k & 4/5 & 1.18 \\
Python & 165k & 24.1 & 80.4M & 70.6-70.8k & 4/5 & 6.78-6.79 \\
C & 2.8M & 2.4 & 547M & 479.6-482.2k & 4/5 & 1.00 \\
Assembly & 2.8M & 20.7 & 1.54B & 1350.8-1353.4k & 4/5 & 0.35 \\
Binary (hex) & 2.8M & 8.1 & 900M & 789.5-792.2k & 4/5 & 0.61 \\
\bottomrule
\end{tabular}
\end{table}

\begin{table}[htbp]
  \centering
  \caption{Corpus diagnostics. All tokens: the full public split before truncation. Train split: training-split tokens after truncation and packing (smallest seed). Passes at 30k: pass equivalents at update 30,000 over the five seeds. Type cov.: share of Pereira stimulus word types present. Peak $r$: maximum over blocks of the seed-42 model on Pereira-243 under the benchmark's own aggregation (median over neuroids), a selected descriptive value. BLiMP: accuracy of the seed-42 model (chance 0.5).}
  \label{tab:corpora}
  \small
  \setlength{\tabcolsep}{4pt}
  \begin{tabular}{lrrrrrrr}
\toprule
Corpus & All tokens & Train split & Passes at 30k & Type cov. & Tok./byte & Peak $r$ & BLiMP \\
\midrule
TinyStories & 463M & 416M & 1.18 & 0.99 & 0.24 & 0.154 & 0.639 \\
Python & 175M & 72.3M & 6.78-6.80 & 0.98 & 0.32 & 0.098 & 0.533 \\
C & 588M & 491M & 1.00 & 0.99 & 0.36 & 0.095 & 0.554 \\
Assembly & 2.22B & 1.38B & 0.35-0.36 & 0.98 & 0.52 & 0.081 & 0.537 \\
Binary (hex) & 1.10B & 808M & 0.61 & 0.01 & 0.44 & 0.085 & 0.528 \\
\bottomrule
\end{tabular}
\end{table}

\section{Neural benchmarks and the encoding pipeline}
\label{app:benchmarks}

Table~\ref{tab:benchmark-table} describes the data as loaded by the Brain-Score language commit \texttt{6650678}; counts come from \path{scripts/reproduction/describe_benchmarks.py}. The loaded data differ from the original studies in ways that matter for interpretation. Pereira2018 \citep{pereira2018universal} recorded experiment 2 (384 sentences in 96 passages from 24 topics) in 8 participants and experiment 3 (243 sentences in 72 passages from 24 topics) in 6; the loader contains 9 participants for the 384-sentence set and groups the 243-sentence set into 24 context groups, so a model's cumulative context spans a topic of about ten sentences. Tuckute2024 \citep{tuckute2024driving} averaged the left-hemisphere language-network responses of five participants to 1,000 six-word sentences; the benchmark provides no ceiling for the single averaged response. Blank2014 \citep{blank2014dissociation} recorded ten participants listening to stories; the loader contains 5 participants, 8 stories, and 12 functional regions per participant, sampled at 2-s volumes, so its 1,317 units are time points with their cumulative story context, not independent sentences. Fedorenko2016 \citep{fedorenko2016ecog} recorded high-gamma ECoG in six patients reading eight-word sentences word by word; the loader contains 5 patients, 97 language-responsive electrodes, and 416 word positions.

\begin{table}[htbp]
  \centering
  \caption{Benchmarks as loaded by the pinned implementation. Units: presentations scored (stimuli or time points). Targets: response dimensions (voxels, functional regions, electrodes, or one averaged response). Subj.: participants contributing to the loaded data (averaged for Tuckute2024).}
  \label{tab:benchmark-table}
  \scriptsize
  \setlength{\tabcolsep}{3.5pt}
  \begin{tabular}{lllrrrl}
\toprule
Benchmark & Modality, task & Unit & Units & Targets & Subj. & Model context \\
\midrule
Pereira-243 & fMRI, sentence reading & sentence & 243 & 8{,}031 & 6 & topic-cumulative \\
Pereira-384 & fMRI, sentence reading & sentence & 384 & 12{,}155 & 9 & passage-cumulative \\
Tuckute2024 & fMRI, sentence reading & sentence & 1{,}000 & 1 & 5 (avg.) & isolated sentence \\
Blank2014 & fMRI, story listening & fMRI time point & 1{,}317 & 60 & 5 & story-cumulative \\
Fedorenko2016 & ECoG, sentence reading & word & 416 & 97 & 5 & sentence-cumulative \\
\bottomrule
\end{tabular}
\end{table}

\paragraph{Encoding.} For each passage, story, or sentence, the benchmark presents stimuli in order and gives the model the space-joined text of all stimuli so far; the representation is the final token's activation at the chosen block, with the model in evaluation mode and gradients disabled. The tokenizer truncates contexts longer than 1,024 tokens from the left, keeping the most recent tokens. Within each training fold, we fit a ridge regression (generalized cross-validation over seven logarithmically spaced penalties $\alpha \in \{10^{-3}, \ldots, 10^{3}\}$, with one $\alpha$ shared across targets, fitted with an intercept and without feature or target standardization). Folds are grouped by context for Pereira (10 folds), by story for Blank2014 (8), and by sentence for Fedorenko2016 (10), and are random 90/10 splits of sentences for Tuckute2024 (10). The benchmark score takes the median over neuroids of the fold-averaged correlation; our scores average the correlation over neuroids within each fold, then over folds, preserving the fold structure for every model. The two aggregations differ by 0.005-0.010 for the seed-42 models; Table~\ref{tab:corpora} and Table~\ref{tab:embed-adapt} report values using the benchmark's own aggregation. Benchmark ceilings use that aggregation and are not comparable with our scores, so we do not normalize by them.

\paragraph{Blank2014 truncation.} Of the 1,317 cumulative contexts, 417 exceed 1,024 tokens (maximum 1,793). The encoding path truncates from the left, so these contexts lose the story's start but remain distinct. Our geometry extraction truncated from the right, keeping the first 1,024 tokens, making 411 contexts byte-identical to an earlier one; therefore, Blank2014 RSA is near zero for every model (0.003-0.013) and is not interpreted. The released scoring script for seeds 43-46 (\path{scripts/score_local_checkpoints.py}) and the fold-score extractor both load the tokenizer with left truncation.

\section{Primary inference and depth-resolved replication}
\label{app:primary}

\begin{figure}[htbp]
  \centering
  \includegraphics[width=0.7\textwidth]{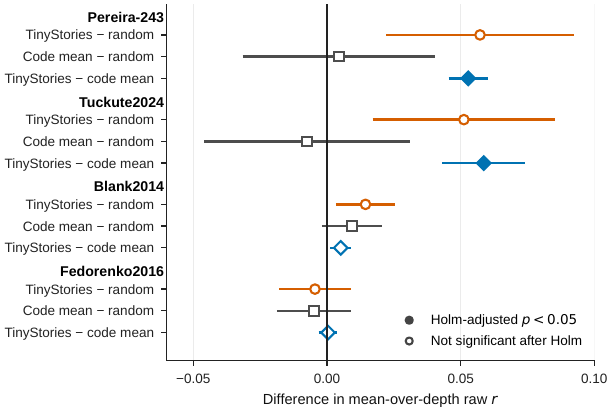}
  \caption{The twelve primary contrasts (Table~\ref{tab:primary-tests}). Lines are unadjusted 95\% intervals; filled markers survive Holm correction over the twelve tests, so an interval excluding zero does not by itself imply a corrected decision. Zero means no change relative to the comparison condition, not zero alignment.}
  \label{fig:primary-forest}
\end{figure}

Table~\ref{tab:primary-tests} gives the complete family of twelve tests. Per-initialization summaries are in the results archive, and the contrasts against random weights include their variance, the appropriate test of a training increment over a sampled random-weight population. No contrast against random weights survives Holm correction. Figure~\ref{fig:primary-forest} shows the same family.

\begin{table}[htbp]
  \centering
  \caption{The twelve primary contrasts on mean-over-depth scores. L: TinyStories; C: code mean; R: random weights. Intervals are unadjusted 95\% Welch intervals; Holm correction covers all twelve rows, and bold marks Holm $p < 0.05$.}
  \label{tab:primary-tests}
  \scriptsize
  \setlength{\tabcolsep}{4pt}
  \begin{tabular}{llrcrrr}
\toprule
Benchmark & Contrast & Estimate & 95\% CI & df & $p$ & Holm $p$ \\
\midrule
Pereira-243 & L-R & +0.057 & [0.022, 0.092] & 2.3 & 0.018 & 0.145 \\
 & C-R & +0.004 & [$-$0.032, 0.040] & 2.2 & 0.675 & 1.000 \\
 & L-C & +0.053 & [0.046, 0.060] & 9.9 & $1.89\times10^{-8}$ & \textbf{$2.26\times10^{-7}$} \\
\addlinespace
Tuckute2024 & L-R & +0.051 & [0.017, 0.085] & 3.4 & 0.016 & 0.145 \\
 & C-R & $-$0.007 & [$-$0.046, 0.031] & 2.4 & 0.534 & 1.000 \\
 & L-C & +0.059 & [0.043, 0.074] & 6.2 & $7.51\times10^{-5}$ & \textbf{$8.26\times10^{-4}$} \\
\addlinespace
Blank2014 & L-R & +0.014 & [0.003, 0.025] & 2.7 & 0.027 & 0.186 \\
 & C-R & +0.009 & [$-$0.002, 0.020] & 2.6 & 0.076 & 0.453 \\
 & L-C & +0.005 & [0.001, 0.009] & 11.1 & 0.013 & 0.134 \\
\addlinespace
Fedorenko2016 & L-R & $-$0.004 & [$-$0.018, 0.009] & 2.5 & 0.330 & 1.000 \\
 & C-R & $-$0.005 & [$-$0.019, 0.009] & 2.3 & 0.297 & 1.000 \\
 & L-C & +0.000 & [$-$0.003, 0.004] & 9.4 & 0.809 & 1.000 \\
\bottomrule
\end{tabular}
\end{table}

\begin{figure}[htbp]
  \centering
  \includegraphics[width=\textwidth]{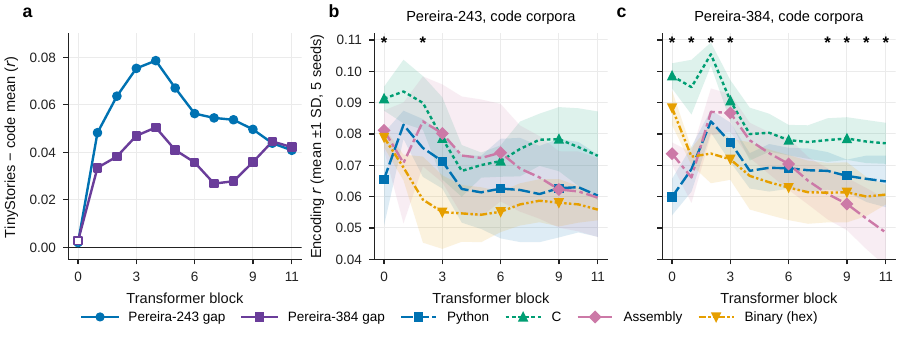}
  \caption{Depth-resolved replication. \textbf{(a)} TinyStories minus the code mean at each block on both Pereira sets; filled markers are significant after Holm correction over blocks (Welch $t$, 5 TinyStories versus 20 code models). \textbf{(b, c)} Code corpora: mean $\pm1$ SD over five seeds; asterisks mark blocks with significant heterogeneity among the four code corpora (one-way ANOVA, Holm over blocks).}
  \label{fig:seed-depth}
\end{figure}

\begin{figure}[htbp]
  \centering
  \includegraphics[width=\textwidth]{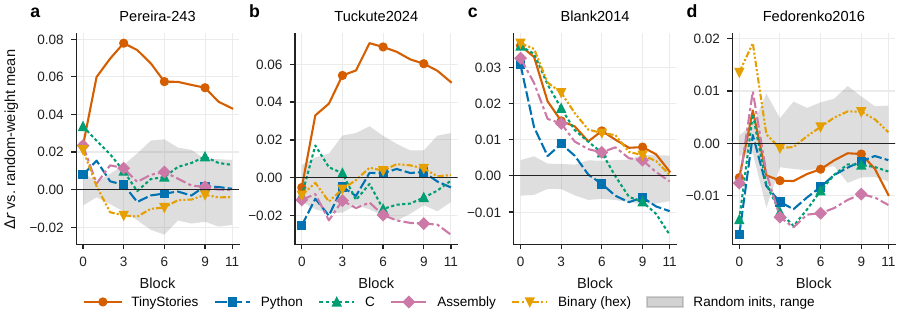}
  \caption{Change relative to the random-weight mean at each block on the four benchmark families (raw $r$; mean of five seeds per corpus). The gray band spans the three random initializations around their mean. Vertical scales differ between benchmarks.}
  \label{fig:all-benchmarks}
\end{figure}

On Pereira-243, the TinyStories-code gap is significant at 11 of 12 blocks after correction over blocks (all but block 0), with a selected maximum of $\Delta r = 0.079$ at block 4; on Pereira-384 it is significant at 11 of 12 blocks (again all but block 0), maximum 0.050 at block 4, mean 0.036; Tuckute2024 also shows 11 of 12. Code heterogeneity is significant at blocks 0 and 2 of Pereira-243 and at 8 of 12 Pereira-384 blocks (Figure~\ref{fig:seed-depth}). Figure~\ref{fig:all-benchmarks} shows each corpus relative to the random-weight mean on all four benchmark families.

\section{Exposure diagnostics and text manipulations}
\label{app:confounds}

\begin{figure}[htbp]
  \centering
  \includegraphics[width=\textwidth]{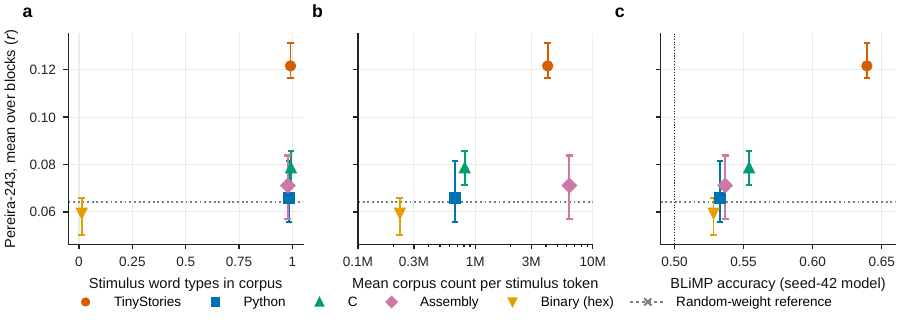}
  \caption{Corpus diagnostics against the Pereira-243 mean over blocks (points: mean of five seeds; error bars: range over seeds). \textbf{(a)} Share of stimulus word types present in the corpus. \textbf{(b)} Mean corpus count per stimulus token (log scale). \textbf{(c)} BLiMP accuracy of the seed-42 model; the dotted vertical line is chance. The dashed line is the matched random-weight reference (mean of three initializations).}
  \label{fig:confounds}
\end{figure}

Figure~\ref{fig:confounds} plots corpus diagnostics against the Pereira-243 mean-over-depth score. Coverage counts Pereira word types that occur at least once in the full corpus. Stimulus-token frequency is the average, over every token occurrence in the Pereira stimuli, of that token's count in the full corpus. The five-corpus regression of the selected seed-42 peak on log frequency gives $r = 0.38$ ($p = 0.53$); with five points it has little power. BLiMP accuracies are from the seed-42 models.

\paragraph{Pseudoword and factorial controls.} Both controls train GPT-2 small on rebuilt TinyStories corpora for 500M tokens with three seeds per condition, checkpointing at 10M, 50M, 100M, 250M and 500M non-padding tokens and scoring blocks 0, 6 and 11 on Pereira-243 with the benchmark's own aggregation. Unlike the primary runs, they do not use early stopping. The form control (condition C) maps each content-word type to one pseudoword. The factorial crosses word order (intact or shuffled within sentences) with lexical identity (intact or each content-word occurrence redrawn from the occurrence pool of its part-of-speech class, preserving unigram frequencies and part-of-speech sequences); the intact cell is shared with the form control.

\begin{figure}[htbp]
  \centering
  \includegraphics[width=\textwidth]{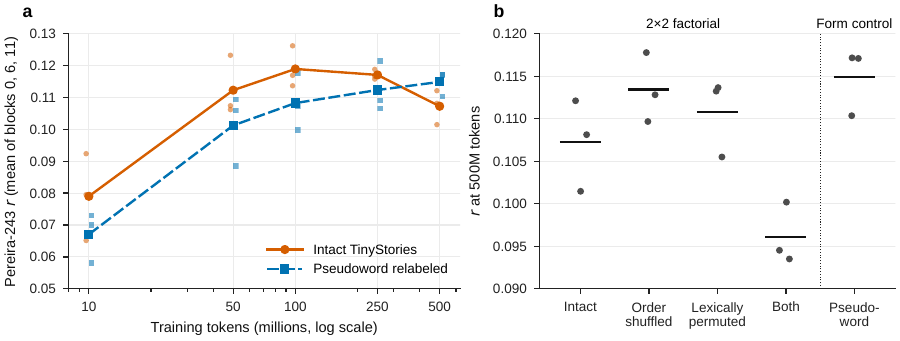}
  \caption{Text manipulations, GPT-2 small, Pereira-243, mean of blocks 0, 6 and 11. \textbf{(a)} Intact versus pseudoword-relabeled TinyStories during training; small markers are seeds (offset horizontally for visibility), lines join seed means. \textbf{(b)} Endpoint at 500M tokens for the four factorial cells and the form control; points are seeds and bars are means. Three seeds per condition; descriptive.}
  \label{fig:factorial}
\end{figure}

\section{Intervention details}
\label{app:interventions}

\begin{figure}[htbp]
  \centering
  \includegraphics[width=\textwidth]{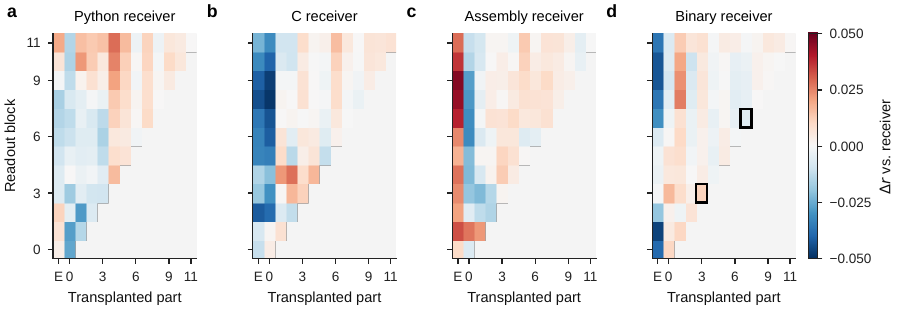}
  \caption{Single-part transplants from the TinyStories donor into each code receiver (Pereira-243). Cells show the change in $r$ for each readout block (rows) after replacing the embeddings (E) or a single block (columns); gray cells appear after the readout and cannot be changed. Outlined cells differ from the receiver after Holm correction over patch positions, using fold-paired tests that are descriptive rather than seed-level inference.}
  \label{fig:single-block}
\end{figure}

\paragraph{Transplants.} Progressive transplants copy the donor's blocks $0,\ldots,k-1$ into the receiver; the second arm also copies the token and position embeddings, re-ties the output head, and copies the final normalization only at $k = 12$. Parameter hashes confirm that $k = 0$ reproduces each receiver and that $k = 12$ with embeddings reproduces the donor. Single-block transplants (Figure~\ref{fig:single-block}) replace a single block or only the embeddings, without re-adapting the normalization. Of the 360 cells in which the transplanted part precedes the readout, two differ from the receiver after Holm correction across patch positions (paired over the 10 benchmark folds), both in the binary receiver and in opposite directions. Because a foreign block meets a receiver's coordinates and normalization statistics, this near-null cannot show that single blocks carry nothing.

\paragraph{Embedding-only adaptation.} Only the token embedding matrix requires gradients; position embeddings, blocks and final normalization are frozen. Because GPT-2 ties this matrix to the output layer, the update changes both the input embedding and the output projection. The launcher configuration uses 2,000 updates of 32 sequences of 64 tokens drawn from a 2M-token sample of the training corpus; the runs' own training logs were not retained. Table~\ref{tab:embed-adapt} reports each model's selected maximum under the benchmark's own aggregation and the prespecified mean over depth.

\begin{table}[htbp]
  \centering
  \caption{Embedding-only adaptation. Peak: each model's maximum over blocks under the benchmark's aggregation (selected, descriptive; the receiver's and the adapted model's peaks can occur at different blocks). Depth: mean over twelve blocks. $\Delta$: change from the untouched receiver. Paired $t$ tests compare English and own-corpus changes across the four receivers.}
  \label{tab:embed-adapt}
  \small
  \setlength{\tabcolsep}{4pt}
  \begin{tabular}{lrrrrrr}
\toprule
Receiver & Peak $r$ & Own $\Delta$ & English $\Delta$ & Depth $r$ & Own $\Delta$ & English $\Delta$ \\
\midrule
Python & 0.098 & +0.000 & +0.014 & 0.066 & +0.002 & +0.031 \\
C & 0.095 & +0.005 & +0.018 & 0.072 & +0.008 & +0.017 \\
Assembly & 0.081 & $-$0.010 & +0.025 & 0.057 & $-$0.004 & +0.032 \\
Binary (hex) & 0.085 & +0.001 & +0.033 & 0.063 & +0.001 & +0.031 \\
\addlinespace
Mean &  & $-$0.001 & +0.022 &  & +0.002 & +0.028 \\
Paired $t(3)$ &  &  & 4.06, $p=0.027$ &  &  & 4.31, $p=0.023$ \\
\bottomrule
\end{tabular}
\end{table}

\section{Geometry details}
\label{app:geometry}

The multi-seed geometry uses 25 trained models and two random initializations (a third was not retained in the geometry run). Table~\ref{tab:geometry-summary} summarizes it.  Here, the values are multi-seed and use the same cumulative contexts as the rest of the table.

\begin{table}[htbp]
  \centering
  \caption{Geometry on Pereira-243. Seed CKA: mean CKA between seeds of a corpus over blocks 4-11. Peak RSA: maximum seed-mean Spearman $\rho$ over blocks, and its block. PR: participation ratio (seed mean) at block 0, its maximum over blocks, and block 11.}
  \label{tab:geometry-summary}
  \small
  \setlength{\tabcolsep}{4pt}
  \begin{tabular}{lrrrrrrr}
\toprule
Model & $n$ & Seed CKA & Peak RSA & Block & PR b0 & PR max & PR b11 \\
\midrule
TinyStories & 5 & 0.899 & 0.139 & 6 & 1.7 & 26.0 & 23.9 \\
Python & 5 & 0.714 & 0.064 & 1 & 3.6 & 40.9 & 40.6 \\
C & 5 & 0.811 & 0.070 & 2 & 1.8 & 30.0 & 28.6 \\
Assembly & 5 & 0.687 & 0.065 & 1 & 2.1 & 19.2 & 18.5 \\
Binary (hex) & 5 & 0.859 & 0.057 & 0 & 2.0 & 3.3 & 2.9 \\
Random weights & 2 & 0.674 & 0.072 & 3 & 83.2 & 83.2 & 61.8 \\
\bottomrule
\end{tabular}
\end{table}

\begin{table}[htbp]
  \centering
  \caption{Corpus separation in CKA. First row: mean CKA among TinyStories seeds minus mean CKA between TinyStories and code models. Second row: mean within-corpus CKA minus mean between-corpus CKA over all five corpora.}
  \label{tab:cka-depth}
  \small
  \begin{tabular}{lrrrrr}
\toprule
Block & 0 & 2 & 4 & 8 & 11 \\
\midrule
TinyStories within $-$ TinyStories-to-code & 0.055 & 0.130 & 0.394 & 0.410 & 0.385 \\
All corpora within $-$ between & 0.060 & 0.118 & 0.317 & 0.337 & 0.333 \\
\bottomrule
\end{tabular}
\end{table}

\begin{table}[htbp]
  \centering
  \caption{RSA after removing leading principal components (Pereira-243). Representations are column-centered before removing the components, which lowers the unablated values relative to Table~\ref{tab:geometry-summary}; each value is the mean over seeds of each model's maximum over blocks.}
  \label{tab:pc-ablation}
  \small
  \begin{tabular}{lrrrr}
\toprule
Model & None removed & 1 PC & 2 PCs & 3 PCs \\
\midrule
TinyStories & 0.075 & 0.061 & 0.057 & 0.042 \\
Python & 0.049 & 0.039 & 0.033 & 0.030 \\
C & 0.047 & 0.028 & 0.030 & 0.029 \\
Assembly & 0.048 & 0.030 & 0.027 & 0.023 \\
Binary (hex) & 0.048 & 0.040 & 0.022 & 0.024 \\
Random weights & 0.042 & 0.025 & 0.024 & 0.026 \\
\bottomrule
\end{tabular}
\end{table}

\paragraph{Other benchmarks.} Peak seed-mean RSA on Pereira-384 is 0.096 for binary and 0.090 for TinyStories, both at block 0; on Fedorenko2016 it is 0.29 for binary, 0.20 for TinyStories and 0.13 for random weights. Tuckute2024 has a single averaged response, so we use Euclidean distance for its neural dissimilarities (correlation distance is undefined in one dimension). RSA is near zero for every model and is not directly comparable with correlation-distance RSA on the other datasets; we discuss Blank2014 in Appendix~\ref{app:benchmarks}.

\begin{figure}[htbp]
  \centering
  \includegraphics[width=\textwidth]{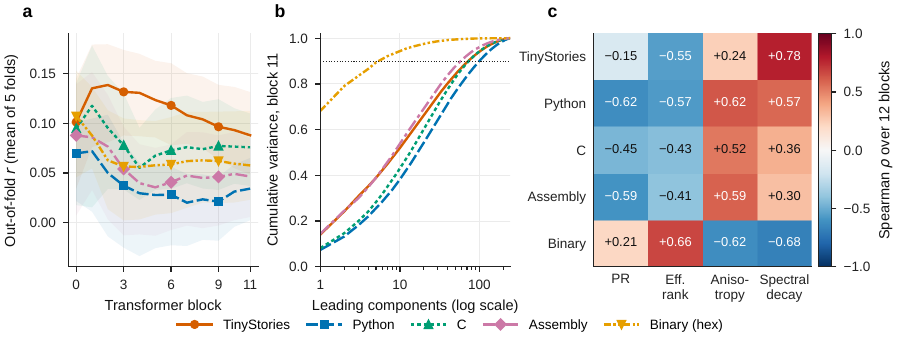}
  \caption{One-checkpoint spectrum audit (seed 42, isolated sentences, five grouped outer folds). \textbf{(a)} Out-of-fold $r$ through depth; lines are fold means and bands span the five folds (descriptive, not a confidence interval). \textbf{(b)} Cumulative variance explained at block 11. \textbf{(c)} Spearman correlation across blocks between four spectral summaries and alignment.}
  \label{fig:geometry-supp}
\end{figure}

\paragraph{One-checkpoint spectrum audit.} Figure~\ref{fig:geometry-supp} uses the seed-42 checkpoints, isolated sentences rather than cumulative contexts, and a separate grouped five-fold ridge encoder averaged over neuroids, so its absolute values are not comparable with the main analyses. At block 11, binary needs 6 components to reach 90\% of variance while the other corpora need 59-100. Its rank correlations across twelve blocks describe how spectrum and alignment co-vary within one model; they are not twelve independent replications.

\section{Inferential scope}
\label{app:scope}

The random-weight models are a matched reference, not a permutation null: they ask whether training adds predictivity beyond the architecture, the tokenizer, and a fitted readout. The TinyStories-code contrast is the primary content comparison because both sides follow the same training and evaluation path. No primary result selects a layer after observing neural scores; selected maxima appear only in labeled descriptive analyses. Seed-level inference generalizes across optimization runs of this design. A hierarchical analysis over seeds, participants, and stimuli would require participant- and item-level outputs that the aggregate scoring interface does not expose.

\FloatBarrier

\section{Broader impacts}
\label{app:impacts}

This is basic research on small models and recorded neural responses. Its practical implication concerns interpretation: absolute brain scores can be over-read when matched references, fit-free analyses, and benchmark-specific outcomes are omitted. The results do not establish how people comprehend computer code; the experiments change a model's training corpus and compare it with human responses to language.

\section{Reproducibility and release}
\label{app:release}

Three levels of reproduction are supported. The released result bundle lets you recompute every reported statistic in minutes on a CPU, without checkpoints or neural data. From checkpoints, the scoring, geometry, and intervention scripts listed in the repository's reproduction guide recompute the bundle; this needs the Brain-Score language submodule, the benchmark data it downloads, and a GPU. Full reproduction retrains from the public corpus with the recorded configurations; comparable 500M-token runs took 10-11 hours of training plus 1.5-6 hours of tokenization on one A40 GPU per model, and results are not bit-identical across devices. The code and result bundle are public at \url{https://github.com/IParraMartin/language-training-brain-geometry}; the seed-42 models and training corpus are public on the Hugging Face Hub. The seed 43-46 checkpoints are not publicly released. Benchmark data are not redistributed and are obtained through the Brain-Score loader.

\end{document}